\documentclass[letterpaper, 10 pt, journal, twoside]{IEEEtran}
\usepackage{hyperref}
\usepackage[T1]{fontenc}
\usepackage{wrapfig}
\usepackage{booktabs}
\usepackage{amsmath}
\usepackage{amssymb}  
\usepackage{accents}
\usepackage{todonotes}
\usepackage{colortbl}
\usepackage{caption}
\usepackage{diagbox}
\usepackage{makecell}
\usepackage[ruled,linesnumbered]{algorithm2e}
\usepackage{subfigure}
\usepackage{float}
\usepackage{placeins}

\ifCLASSINFOpdf
\else
\fi
\begin{document}
%
\title{S$^2$-Diffusion: Generalizing from Instance-level to Category-level Skills in Robot Manipulation}

\author{Quantao Yang$^{*1}$, Michael C. Welle$^{*1,2}$, Danica Kragic$^1$, and Olov Andersson$^1$%
\thanks{Manuscript received: June, 11, 2025; Revised September, 11, 2025; Accepted October, 4, 2025.}
\thanks{This paper was recommended for publication by Editor Aleksandra Faust upon evaluation of the Associate Editor and Reviewers' comments.
This work was supported by Knut and Alice Wallenberg Foundation through Wallenberg AI, Autonomous Systems, and Software Program (WASP) and the European Union's Horizon Europe Framework Programme under grant agreement No 101070596 (euROBIN).} 
\thanks{$*$These authors contributed equally.}
\thanks{$^{1}$
Division of Robotics, Perception and Learning (RPL), KTH Royal Institute of Technology, Sweden. {\tt (e-mail: quantao@kth.se)}.
}
\thanks{$^{2}$Michael C. Welle is also with INCAR Robotics AB, Sweden.}
\thanks{Digital Object Identifier (DOI): see top of this page.}
}

\markboth{IEEE Robotics and Automation Letters. Preprint Version. Accepted OCTOBER, 2025}
{Yang \MakeLowercase{\textit{et al.}}: S$^2$-Diffusion}

\maketitle

\begin{abstract}
Recent advances in skill learning has propelled robot manipulation to new heights by enabling it to learn complex manipulation tasks from a practical number of demonstrations. However, these skills are often limited to the particular action, object, and environment \textit{instances} that are shown in the training data, and have trouble transferring to other instances of the same category.
In this work we present an open-vocabulary Spatial-Semantic Diffusion policy (S$^2$-Diffusion) which enables generalization from instance-level training data to category-level, enabling skills to be transferable between instances of the same category. We show that functional aspects of skills can be captured via a promptable semantic module combined with a spatial representation. We further propose leveraging depth estimation networks to allow the use of only a single RGB camera.
Our approach is evaluated and compared on a diverse number of robot manipulation tasks, both in simulation and in the real world. Our results show that S$^2$-Diffusion is invariant to changes in category-irrelevant factors as well as enables satisfying performance on other instances within the same category, even if it was not trained on that specific instance. Project website: \url{https://s2-diffusion.github.io}.
\end{abstract}

\begin{IEEEkeywords}
Imitation Learning, Learning from Demonstration, Deep Learning in Grasping and Manipulation.
\end{IEEEkeywords}

%
\IEEEpeerreviewmaketitle

\section{Introduction}
\label{sec:introduction}
%
%
%
%
\IEEEPARstart{I}{mitation} learning (IL)~\cite{mandlekar2020learning, chi2023diffusionpolicy} has shown potential in enabling robotic manipulation in challenging real-world scenarios by learning complex skills from human demonstrations. Still, existing IL methods often struggle to generalize beyond the specific training environments from which the demonstrations are derived. This is an important obstacle as each new environment requires labor-intensive data collection, model fine-tuning, and retraining to adapt the learned policies.

For humans, transferring knowledge between tasks and skills, such as transferring the scooping skill from rice to cereals, is rather straightforward. Scooping rice or cereals may be considered as different instances of the same task for current IL methods. 
The ability to generalize over such instances is still a challenge and requires rather advanced spatial-semantic understanding~\cite{zhu2023viola}.
The ability to transfer and generalize over instances removes the necessity for extensive training and also allows for assessing what type of instances one can transfer over - for example, scooping ice cream may be very different from scooping granular material such as cereals or rice. Thus, granular materials may be seen as the same category as rice and cereals, while ice cream is an instance of another category which requires a very different policy when executing the scooping task. 

\begin{figure}[t]
	\centering
	\includegraphics[width=0.9\linewidth]{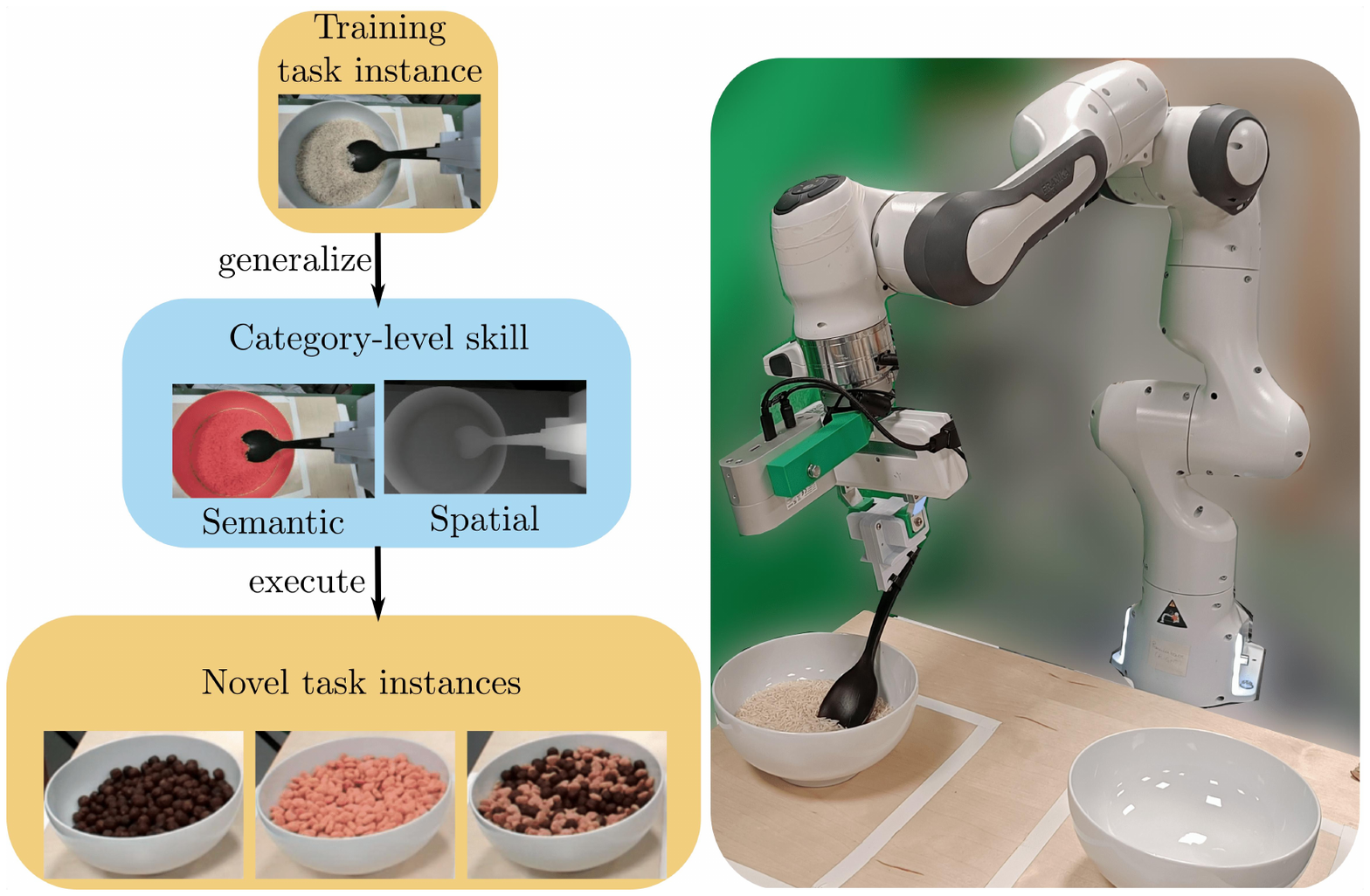}
	\vspace{-0.1cm}
	\caption{Our Spatial-Semantic Diffusion policy (S$^2$-Diffusion) not only efficiently completes the task at hand but also enables the generalization of the same skill across diverse contexts and task variations. 
    }
	\label{fig:frontpage} 
	\vspace{-0.0cm}
    \vspace{-\baselineskip}
    \vspace{-\baselineskip}
\end{figure}

Large pre-trained Visual-Language-Action (VLA) models~\cite{o2023open, team2024octo, wen2024tinyvla} generalize simple skills such as pick-and-place over a wide range of environments and objects. However, more complex non-prehensile manipulation such as scooping, mug-flipping, cooking shrimp or opening a bottle with a bottle opener stay elusive for such large general models.
Recent imitation learning approaches address learning of instances of challenging manipulation tasks ~\cite{chi2023diffusionpolicy, zhao2023learning, gao2024prime, fan2025diffusion, ingelhag2024robotic} but the integration with semantic knowledge in highly-complex manipulation tasks remains a challenge. The aforementioned methods
often rely on raw perceptual features and environmental conditions, limiting their applicability 
to the instances observed during the training. Training a skill via imitation learning, such as a diffusion policy~\cite{chi2023diffusionpolicy}, depends on expert demonstrations that often do not cover several instances of the same task. We show 
that when trained only on a particular instance of a task - such as wiping red scribbles from a whiteboard - the skill fails to transfer already when the scribbles are now green, even if the required action and environment for wiping is exactly the same. This is because the policy did not learn a \textit{whiteboard-wiping} category skill but a single instance of this category namely \textit{red-whiteboard-wiping}.

Motivated by the above challenges, we present a novel approach that integrates spatial prediction with semantic segmentation features from large pre-trained models~\cite{liang2023open, yang2024depth} to generalize from expert demonstrations on a single instance of a task to its \textit{category-level skill} -- such as wiping different color scribbles from the whiteboard or scooping different granular material from bowl to bowl. 
Our method uses a single RGB camera view combined with the proprioceptive information of the robot.
As shown in Fig.~\ref{fig:frontpage}, we extract semantic information using the prompted foundation model~\cite{ren2024grounded} and combine it with a depth estimation foundation model~\cite{yang2024depth2} to obtain \textit{Spatial-Semantic} observations for the visuomotor diffusion policy learning framework.
This allows for invariance to task-irrelevant factors such as background and object textures
as well as the capability to generalize from instance-level to category-level skills. 
Our key contributions are threefold:
\emph{i)} We propose to endow visuomotor diffusion policies with spatial-semantic understanding to enable generalization from instance-level to category-level skills in robot manipulation tasks.
\emph{ii)} We introduce an efficient representation of the spatial-semantic information via a combination of vision foundation models. The overall framework is real-time viable and requires only a single RGB camera and the robot's proprioceptive observations.
\emph{iii)} Our extensive experiments evaluate the method on a set of robotics manipulation tasks in simulation and the real world, demonstrating the ability to learn generalizable and effective robotic manipulation policies.
All real-world experimental videos ($246$) are on the project website.

%

\section{\uppercase{Related Work}}
\label{sec:related_work}

\noindent
\textbf{Visual Feature Based Imitation Learning.}
Visual imitation learning methods~\cite{mees2022matters, ha2023scaling} have shown strong potential in diverse robot manipulation tasks. While Zhu et al.~\cite{zhu2023viola} and Wu et al.~\cite{wu2023unleashing} highlight the benefits of object-centric and large-scale video pretraining, they do not demonstrate generalization to novel task instances. 3D vision-based methods~\cite{ke20243d,  gervet2023act3d,  qiu2024learning} improve generalization but require costly multi-view RGB-D setups or scene scanning. In contrast, our method uses a semantic-aware representation from a single RGB image. A concurrent work, SAM2Act~\cite{fang2025sam2act}, introduces a multi-view transformer-based robotics policy for improved visual feature representation and multitask generalization, with SAM2Act+ further incorporating a memory-based architecture for episodic recall in spatial memory-dependent manipulation tasks. Closest to our work, Wang et al.~\cite{wang2024gendp} integrate explicit spatial-semantic information into policy learning, but rely on multi-view 3D descriptor fields, whereas we achieve similar benefits without extra hardware. 

\noindent
\textbf{Vision-Language Models in Robotics.}
Vision-Language Models (VLMs) trained on large internet corpora~\cite{radford2021learning, shang2024theia} have become prominent in robotics. Recent works~\cite{liu2024ok,  wang2024gendp} leverage 2D foundation models like CLIP~\cite{radford2021learning} and SAM~\cite{kirillov2023segment} to construct open-vocabulary 3D representations. Newer models~\cite{liang2023open, zou2024segment} improve open-world generalization. VLMs are also employed to generate high-level task plans~\cite{ahn2022can, chen2023open} or reward functions~\cite{ma2023eureka, mahmoudieh2022zero}. Reuss et al.~\cite{reuss2024multimodal} introduce multimodal diffusion policies for long-horizon tasks. MOKA~\cite{liu2024moka} uses keypoint affordances for manipulation, but we enhance generalization further by incorporating both semantic and depth information.

\noindent
\textbf{Vision-Language Actions in Robotics.}
Vision-Language Actions (VLAs) pretrained on large-scale data~\cite{brohan2023rt, team2024octo, kim2024openvla} show strong task reasoning but remain computationally intensive and struggle with fine-grained manipulation. Cross-embodied policies~\cite{doshi2024scaling} attempt to scale generalization across robots. Instead, our method focuses on compact, efficient policies by leveraging open-vocabulary segmentation masks for policy generalization across manipulation tasks.

\begin{figure}
    \centering
    \vspace{0.3cm}
    \includegraphics[width=0.99\linewidth]{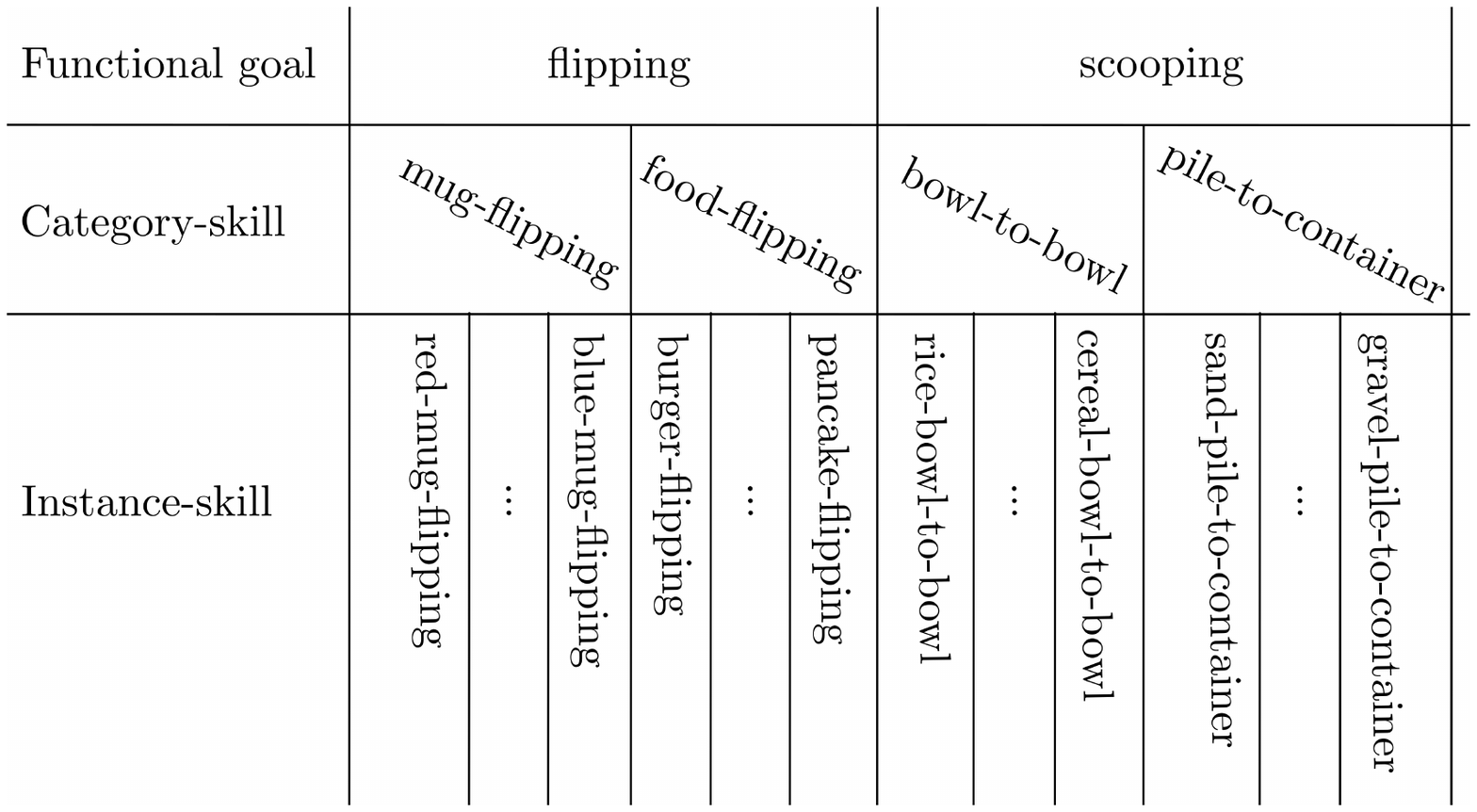}
    \caption{Skill abstraction hierarchy for flipping and scooping tasks.}
    \label{fig:skill_h}
    \vspace{-\baselineskip}
\end{figure}
\section{Towards Generalizable Robotic Skills}

What does it mean for a skill to be \textit{generalizable}? Commonly, skills like "mug flipping" learned on a specific instance do not transfer to new instances with different appearance or geometry. Neural policies often overfit to instance-specific features, limiting their applicability.
One solution is to explicitly learn features shared across a category. For instance, "mugs" can be characterized by a cylindrical body and handle. Datasets like ShapeNet~\cite{chang2015shapenet} facilitate learning such category-level features.
Object-centric representations work well for defined object classes (e.g., mugs, shoes), however, they struggle with tasks involving materials or actions, such as wiping or scooping. 
Instead, we advocate an \textit{action-centric} perspective, focusing on \textit{semantic functionality} over object identity.

\begin{figure*}[!t]
    \centering
    \includegraphics[width=1.0\linewidth]{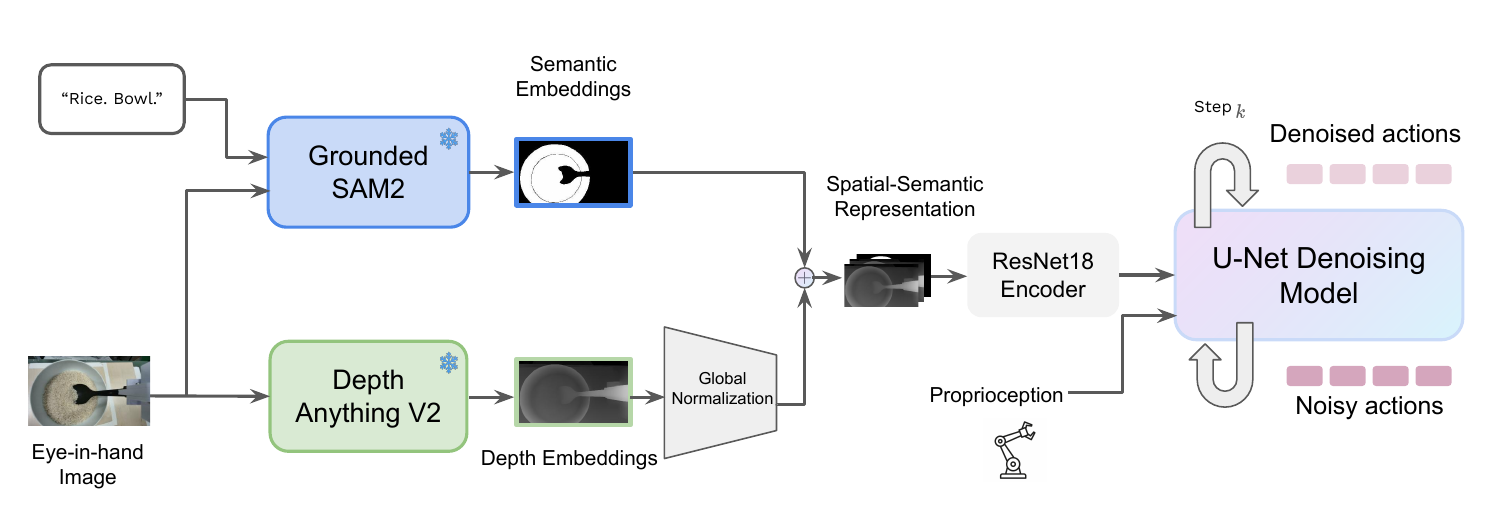}
    \vspace{-0.5cm}
      \caption{\textbf{S$^2$-Diffusion Architecture}. The architecture is composed of three components: a pretrained semantic segmentation model \textit{Grounded-SAM2}~\cite{ren2024grounded}, a pretrained depth prediction model \textit{DepthAnythingV2}~\cite{yang2024depth} and a U-Net denoising diffusion policy~\cite{chi2023diffusionpolicy}. We design an object-aware spatial-semantic representation that is leveraged for denoising probabilistic model.}
      \vspace{0mm}
      \label{fig: architecture}
      \vspace{-\baselineskip}
\end{figure*}

Skills should be abstracted beyond specific instances. As illustrated in Fig.~\ref{fig:skill_h}, a functional goal like scooping encompasses various category-level skills and their specific instances. For example,
scooping represents transferring granular or semi-solid materials. A category-level skill like bowl-to-bowl scooping covers materials such as rice, cereal, or lentils, whereas instance-level policies (e.g., rice scooping) often fail to generalize.
In real-world robotics, demonstrations naturally reflect instance-level setups. Generalization could be attempted by collecting demonstrations across many instances, but this is costly and labor-intensive. Instead, we propose leveraging Spatial-Semantic features extracted from a single instance. By replacing raw RGB inputs with spatial-semantic observations, we enable visuomotor diffusion policies to generalize from instance-level training to category-level execution.

\section{\uppercase{Problem Formulation}}
Our goal is to learn a generalizable imitation learning policy by leveraging spatial-semantic representation from pretrained Vision-Language models. 
We assume access to a dataset $D=\{\tau_i\}_i^N$ of $N$ demonstrated expert trajectories $\tau_i=\{(o_0,a_0),...,(o_{T_i}, a_{T_i})\}$ for the task. Our method uses a denoising diffusion process~\cite{chi2023diffusionpolicy} to learn imitation policy $\pi$ from $D$. To generalize
from individual instances to unseen instances from the same category,
we propose to build a spatial-semantic representation. 
$\mathcal{O}$ is the observation space composed of visual spatial-semantic representation $f_v$ and
robot proprioception states $q$. $\mathcal{A}$ is the action space of robot end-effector commands. We aim to learn a policy $\pi_{\theta}(a|o): \mathcal{O} \rightarrow \mathcal{A}$ with parameters $\theta$ that predicts action $a$ according to current observation $o$ by leveraging the prior experience contained in the dataset $D$.

\section{\uppercase{Method}}
\label{sec:method}

Our objective is to develop an open-vocabulary spatial-semantic visuomotor policy that can generalize from an individual instance to other unseen instances resulting in a category-level skill.
We propose open-vocabulary Spatial-Semantic Diffusion policy (S$^2$-Diffusion), an approach that leverages three main components in policy learning: a semantic segmentation model, a depth prediction model and a diffusion policy shown in Fig.~\ref{fig: architecture}. The policy is trained with demonstrations from expert teleoperation, using only RGB images and robot proprioception as the state, and end-effetor velocities as the commanded actions respectively.

\subsection{Diffusion for Robot Skill Learning}
Following previous works~\cite{chi2023diffusionpolicy, Ze2024DP3}, we formulate the visuomotor policy as a conditional Denoising Diffusion Probabilistic Model (DDPM)~\cite{ho2020denoising, pearce2023imitating}. Starting from random action $a^K$ sampled from Gaussian noise, the diffusion probabilistic model $\epsilon_{\theta}$ performs $K$ iterations of denoising. This process gradually produces a sequence of actions with decreasing noise levels, $a^K, a^{k-1}, ..., a^0$, until the noise-free action $a^0$. Each action denoising iteration is described as: 
\begin{equation}
    a^{k-1} = \alpha_k(a_k-\gamma_k\epsilon_{\theta}(a^k, o, k))+\sigma_k\mathcal{N}(0, I),\\
\end{equation}
where $o$ is the observation for the policy. $\alpha_k$, $\gamma_k$ and $\sigma_k$ are referred as noise schedule for each iteration $k$, and $\mathcal{N}(0, I)$ is the Gaussian noise added to the action.

To learn the action predicting model $\pi_{\theta}$, we randomly sample the robot action $a^0$ from the demonstration dataset $D$ and add the noise $\epsilon^k$ for a random iteration $k$. The training loss for the diffusion model is formulated as:
\begin{equation}
\label{eq: training_loss_1}
    \mathcal{L} = \textit{MSE}(a^0, \pi_{\theta}(a^0+\epsilon^k, o, k)), \\
\end{equation}
where we use an action sampling approach rather than a noise prediction model to enhance the generation of high-dimensional actions.

\begin{algorithm}
    \SetKwInOut{Input}{Input}
    \SetKwInOut{Output}{Output}
    \vspace{-0.0cm}
    \Input{Semantic query $l$, image observation $o_t$, robot state $q$, \textit{Grounded-SAM2} model $C_1$, \textit{DepthAnythingV2} model $C_2$ 
    }
    Collect demonstrated trajectories $D=\{\tau_i\}_i^N$\\
    \For{epoch n=1,N}
    {          
        Sample raw image $o_t$ and robot action $a^0$ \\
        Obtain spatial-semantic representation \\
        \hspace{1cm} $z_s=C_1(o_t, l)$, $z_d=C_2(o_t)$ \\
        \hspace{1cm} $z_f(i, j) = \max_{s=1}^{n} z_s(i, j)$ \\
        \hspace{1cm} $z = z_f \oplus  z_d$ \\
        \hspace{1cm} $f_v = \text{ResNet}(z)$ \\
        
        Add Gaussian noise $\epsilon^k \sim \mathcal{N}(0, I)$ for step $k$ \\
        \hspace{1cm} $a^k=a^0+\epsilon^k$ \\
        
        Train the policy \\
        \hspace{1cm} $\mathcal{L} = \textit{MSE}(a^0, \pi_\theta (a^k, k, f_v, q))$
    }    
    \Return the trained policy $\pi_\theta(a_t|o_t, q, k)$
    \caption{Learning Open-Vocabulary Spatial-Semantic Diffusion Policy}
    \label{alg:training policy}
\end{algorithm}

\subsection{Open-Vocabulary Spatial-Semantic Representation}
\label{sec:ov representation}
We aim to design spatial-semantic-aware representation that is leveraged for the observation of the above denoising probabilistic model. We utilize two pretrained VLMs, \textit{Grounded-SAM2}~\cite{ren2024grounded} and \textit{DepthAnythingV2}~\cite{yang2024depth}, for open-vocabulary semantic segmentation and depth map estimation respectively. We use \textit{Grounded-SAM2} model to perform zero-shot semantic segmentation, leveraging CLIP-based~\cite{radford2021learning} mask classification to segment unseen classes. We combine the extracted features to construct a spatial-semantic representation that is leveraged as the input for the visuomotor diffusion policy. 
We utilize \textit{Grounded-SAM2} to segment an image into a set of semantic masks $(z_1, z_2, ..., z_{n})$ from visual observations based on text descriptions. We apply pixel-wise maximum pooling for each pixel location $(i, j)$ across all segmentation masks:
\begin{equation}
z_f(i, j) = \max_{s=1}^{n} z_s(i, j),
\end{equation}
where \( z_s(i, j) \) represents the pixel value at position \( (i, j) \) in the \( s \)-th mask. In this way, we combine multiple segmentation masks into a single mask $z_f$ where each pixel represents the most confident prediction from the set of masks.

To improve the spatial reasoning of the imitation policy, we propose to incorporate depth map of the task space into its semantic representation. Specifically, the input image is also processed separately with a pretrained 335M-parameter \textit{
DepthAnythingV2} model~\cite{yang2024depth} that shows promising performance in fine-grained details. The 
model predicts depth estimates $z_d$ relative to the input RGB observations rather than absolute values. This can result in inconsistencies and inaccuracies over extended tasks or manipulation horizons. To address this, we normalize the depth maps during both training and evaluation, ensuring consistency across diverse scenes and improving the model's robustness and reliability.
The resulting semantic and spatial feature vectors are concatenated along the channel dimension to form the spatial-semantic representation $z = z_f \oplus z_d$, where $z \in \mathbb{R}^{(C_s + C_d) \times H \times W}$ with $C_s, C_d$ representing the channel dimensions of the semantic and depth features. Upon generating the spatial-semantic representation, we leverage it as input for a visuomotor diffusion policy~\cite{chi2023diffusionpolicy}.

\begin{figure*}[t!]
    \centering
    \vspace{0.2cm}
    \includegraphics[width=0.95\linewidth]{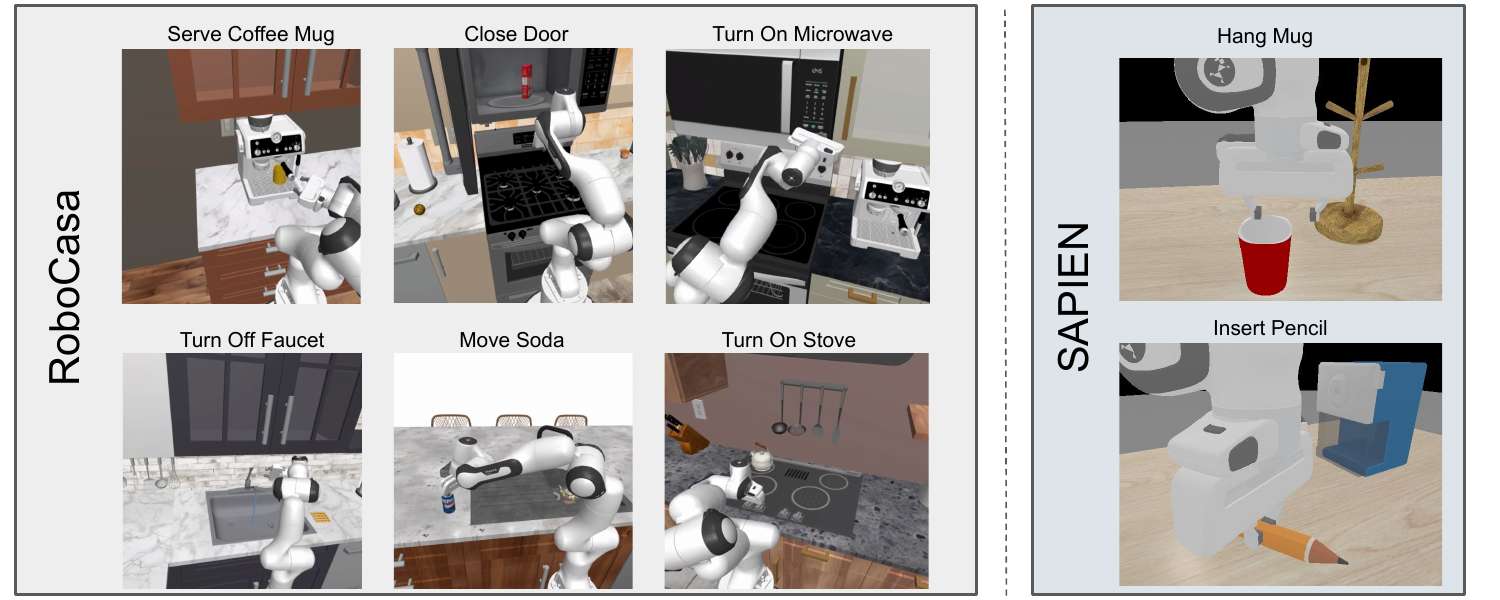}
    \vspace{-0.2cm}
      \caption{\textbf{Simulated Tasks}. We perform evaluations on six single-stage tasks from a large-scale simulation framework RoboCasa~\cite{robocasa2024}: \textit{ServeMug}, \textit{CloseDoor}, \textit{TurnOnMicrowave}, \textit{TurnOffFaucet}, \textit{MoveSoda}, \textit{TurnOnStove}, and two tasks in SAPIEN simulator: \textit{HangMug}, \textit{InsertPencil}. 
      }
      \vspace{0mm}
      \label{fig:sim-tasks}
\end{figure*}

\begin{table*}[ht]
\centering
\caption{Success Rate of Simulation Experiments in RoboCasa
}
\label{table: sim results}
\begin{tabular}{ccccccc}
\toprule
\textbf{Method} & \multicolumn{6}{c}{\textbf{Tasks}}\\
\cmidrule(lr){2-7}
& ServeMug & CloseDoor & TurnOnMicrowave & TurnOffFaucet & MoveSoda & TurnOnStove\\
\midrule
BC-RNN & 0.02 [0.01, 0.07] & 0.03 [0.01, 0.08] & 0.05 [0.02, 0.11] & 0.27 [0.19, 0.36] & 0.06 [0.03, 0.12] & 0.26 [0.18, 0.35] \\
BC-Transformer & 0.16 [0.10, 0.24] & 0.55 [0.45, 0.64] & 0.70 [0.60, 0.78] & 0.31 [0.23, 0.41] & 0.10 [0.06, 0.17] & 0.34 [0.25, 0.44] \\
Diffusion Policy & 0.15 [0.09, 0.23] & 0.52 [0.42, 0.62] & 0.41 [0.32, 0.51] & 0.38 [0.29, 0.48] & 0.22 [0.15, 0.31] & 0.50 [0.40, 0.60] \\
\rowcolor{blue!20} S$^2$-BC-Transformer & 0.54 [0.44, 0.63] & 0.77 [0.68, 0.84] & 0.72 [0.63, 0.80] & 0.55 [0.45, 0.64] & 0.44 [0.35, 0.54] & 0.60 [0.50, 0.69] \\
\rowcolor{gray!30} S$^2$-Diffusion (Ours) & \textbf{0.75 [0.66, 0.82]} & \textbf{0.80 [0.71, 0.87]} & \textbf{0.80 [0.71, 0.87]} & \textbf{0.77 [0.68, 0.84]} & \textbf{0.83 [0.74, 0.89]} & \textbf{0.88 [0.80, 0.93]} \\
\bottomrule
\end{tabular}
\end{table*}

\begin{table}[t]
\centering
\caption{
Simulation Experiments in SAPIEN}
\label{tab:sapen_results}
\setlength{\tabcolsep}{4pt}
\renewcommand{\arraystretch}{1.05}
\resizebox{\columnwidth}{!}{%
\begin{tabular}{lcccc}
\toprule
\textbf{Method} & \multicolumn{2}{c}{\textbf{InsertPencil}} & \multicolumn{2}{c}{\textbf{HangMug}} \\
\cmidrule(lr){2-3} \cmidrule(lr){4-5}
& Seen & Unseen & Seen & Unseen \\
\midrule
GenDP & \textbf{0.95 [0.76, 0.99]} & 0.85 [0.64, 0.95] & \textbf{0.90 [0.70, 0.97]} & 0.80 [0.58, 0.92] \\
\rowcolor{gray!30} S$^2$-Diffusion (Ours) & 0.90 [0.70, 0.97] & \textbf{0.90 [0.70, 0.97]} & 0.85 [0.64, 0.95] & \textbf{0.90 [0.70, 0.97]} \\
\bottomrule
\end{tabular}
}
\vspace{-0.3cm}
\end{table}

\subsection{Learning Semantic Diffusion Policy}
To effectively utilize our open-vocabulary spatial-semantic representations, we adopt the CNN-based diffusion policy architecture~\cite{chi2023diffusionpolicy} as our decision-making backbone. We use Denoising Diffusion Implicit Models (DDIM)~\cite{song2020denoising} as the noise scheduler. In our paper, the observation is composed of spatial-semantic feature $f_v$ and robot proprioceptive state $q$. The training loss in Equation~\ref{eq: training_loss_1} is defined as:
\begin{equation}
    \label{eq: training_loss_2}
   \mathcal{L} = \textit{MSE}(a^0, \pi_{\theta}(\alpha_k a^0+\beta_k \epsilon^k,k,f_v,q)),\\
\end{equation}
where $\alpha_k$ and $\beta_k$ are used for noise schedule of each step. The learning process for our S$^2$-Diffusion method is described in Algorithm~\ref{alg:training policy}. We combine features inferred by VLMs \textit{Grounded-SAM2} and \textit{DepthAnythingV2} to construct the spatial-semantic representation for the action denoising model. By conditioning on the tuple $(a^k, k, f_v, q)$, the denoising model learns to predict the clean action by using mean square error loss for action supervision.

\section{\uppercase{Evaluation}}
\label{sec:method}

The goals of all our experiments are three-fold:
\begin{enumerate}
    \item to evaluate and compare the performance of our method on challenging robotic manipulation tasks;
    \item to validate that generalization from instance-level to category-level skill is achieved in a real-world setting;
    \item to perform ablations in order to investigate the role of the semantic and spatial components.
\end{enumerate}
We first describe the experiments performed in the simulation, followed by the experiments on physical hardware.


\subsection{Simulation Experiments}

\noindent
\textbf{Experiment Setup.}
For our simulation experiments, we take advantage of a recent open-sourced large-scale simulation environment, \textit{RoboCasa}~\cite{robocasa2024}, which provides expert demonstrations for diverse everyday tasks. We evaluate our method 
and the baselines on one task of each atomic task category except navigation and group doors and drawers due to their similarity as shown in
Fig.~\ref{fig:sim-tasks}: \textit{ServeMug}, \textit{CloseDoor}, \textit{TurnOnMicrowave}, \textit{TurnOffFaucet}, \textit{MoveSoda}, and \textit{TurnOnStove} using the provided $50$ expert demonstrations from RoboCasa. The prompts used were: "Mug", "Door", "Microwave button", "Faucet", "Soda can", "Knob" respectively.

\noindent
\textbf{Baselines.} We compare our method with three baseline methods in simulation: 1) \textbf{BC-RNN}: a behavior cloning method with recurrent neural network implementation; 2) \textbf{BC-Transformer}: a common behavior cloning method with transformer architecture~\cite{mandlekar2021matters}; 3) \textbf{Diffusion Policy}: the image-based diffusion policy of \cite{chi2023diffusionpolicy}; 
for completeness we also compare against 4) \textbf{S$^2$-BC-Transformer}: BC-Transformer trained with spatial-semantic data; and 5) \textbf{GenDP}: 
a multi-view RGB-D method~\cite{wang2024gendp}. As RoboCasa does not provide depth, we compare against GenDP on \textit{InsertPencil} and \textit{HangMug} in SAPIEN ~\cite{xiang2020sapien}. 
In all our experiments, we train for $500$ epochs on an NVIDIA RTX 4090 GPU. We set the initial learning rate to $1\text{e}{-4}$, applied a $500$-step linear warm-up, and then used a cosine learning rate scheduler for the remainder of training.

\noindent
\textbf{Simulation Results.}
For simulation experiments in RoboCasa, we take advantage of the open-sourced implementation for the baseline methods from \textit{RoboMimic}~\cite{mandlekar2021matters}. In the demonstration datasets for each task, we assume \textbf{variations} set by RoboCasa in the color of the target object, as well as differences in the background environment. We evaluate $100$ trials with the default seed for each task and the corresponding success rates and the $95\%$ confidence intervals (computed via the Wilson score method for binomial proportions)
are listed in Table~\ref{table: sim results}. The results are
demonstrating matching or statistically significantly superior performance of our S$^2$-Diffusion approach.

Due to the challenging variations in target object and background seen within the RoboCasa dataset, classical imitation learning policies struggle to solve the specific single-stage manipulation task. While the baselines—BC-RNN, BC-Transformer, and Diffusion Policy—show relatively poor performance, particularly on tasks like ServeMug and MoveSoda, S$^2$-Diffusion consistently achieves the highest success rates. This lower performance of both BC and diffusion policy on RoboCasa is likely due to the high diversity of scenes in the expert training data.
We also trained S$^2$-BC-Transformer on the spatial-semantic dataset. We extended BC-Transformer because in our prior evaluation it exhibited comparable performance with DP. While S$^2$-BC-Transformer achieves competitive results on some tasks, S$^2$-diffusion demonstrates more robust performance. Additionally, S$^2$-BC-Transformer outperforms the RGB-only version, highlighting the value of spatial-semantic representations for improving the generalization of imitation policies.
Similar to findings in RoboCasa~\cite{robocasa2024}, the image-based DP is sensitive to in object color and background. In contrast, by leveraging spatial-semantic features our method shows better robustness and generality compared to state-of-the-art alternatives.
We attribute this to the rich visual representation of combining semantic information and depth estimation for the workspace scene.
 Finally, Table~\ref{tab:sapen_results} shows that our RGB method performs on par with the multi-view RGB-D method GenDP.



\begin{figure*}[t]
    \centering
    \subfigure[] {
        \includegraphics[width = 0.42\linewidth]{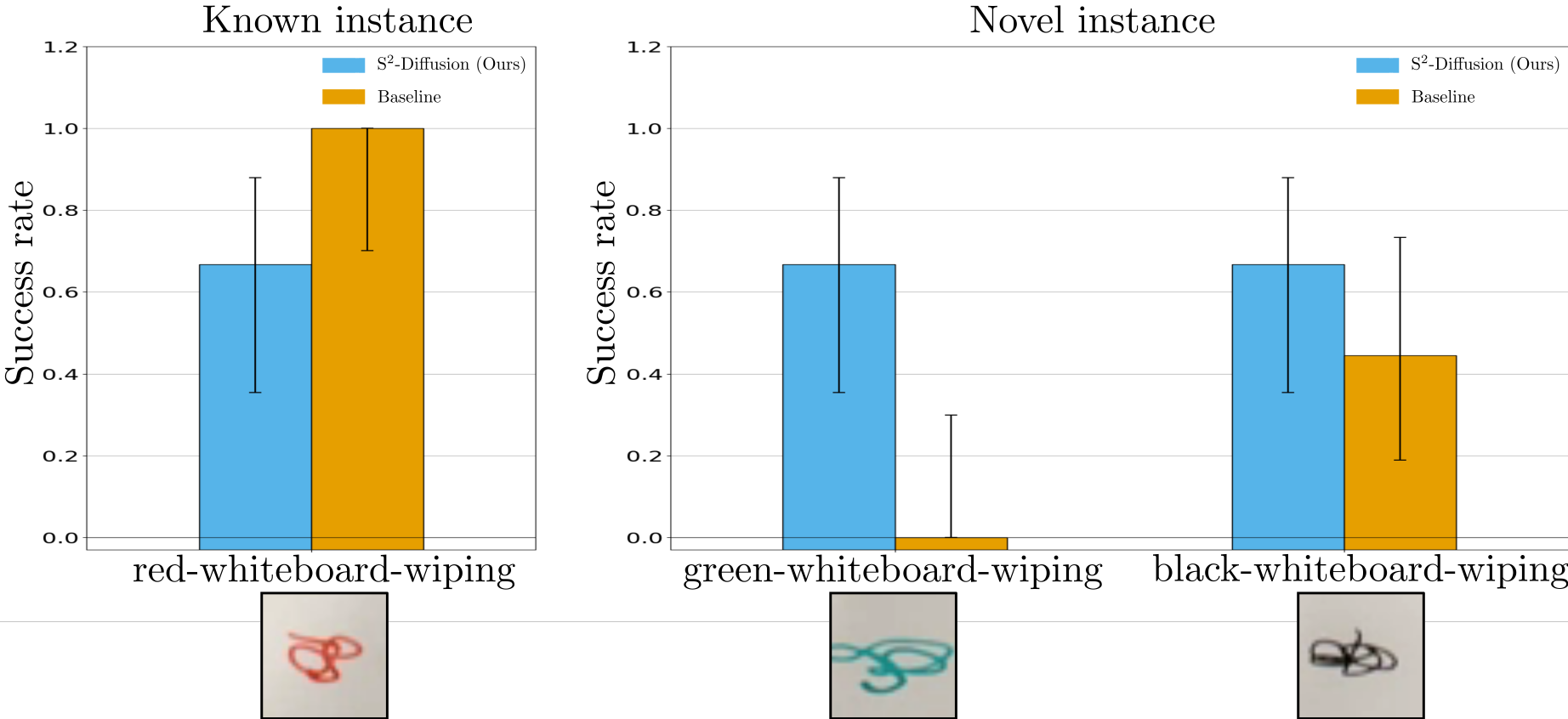}
        \label{fig:real wiping}
    }
    \subfigure[] {
        \includegraphics[width = 0.50\linewidth]{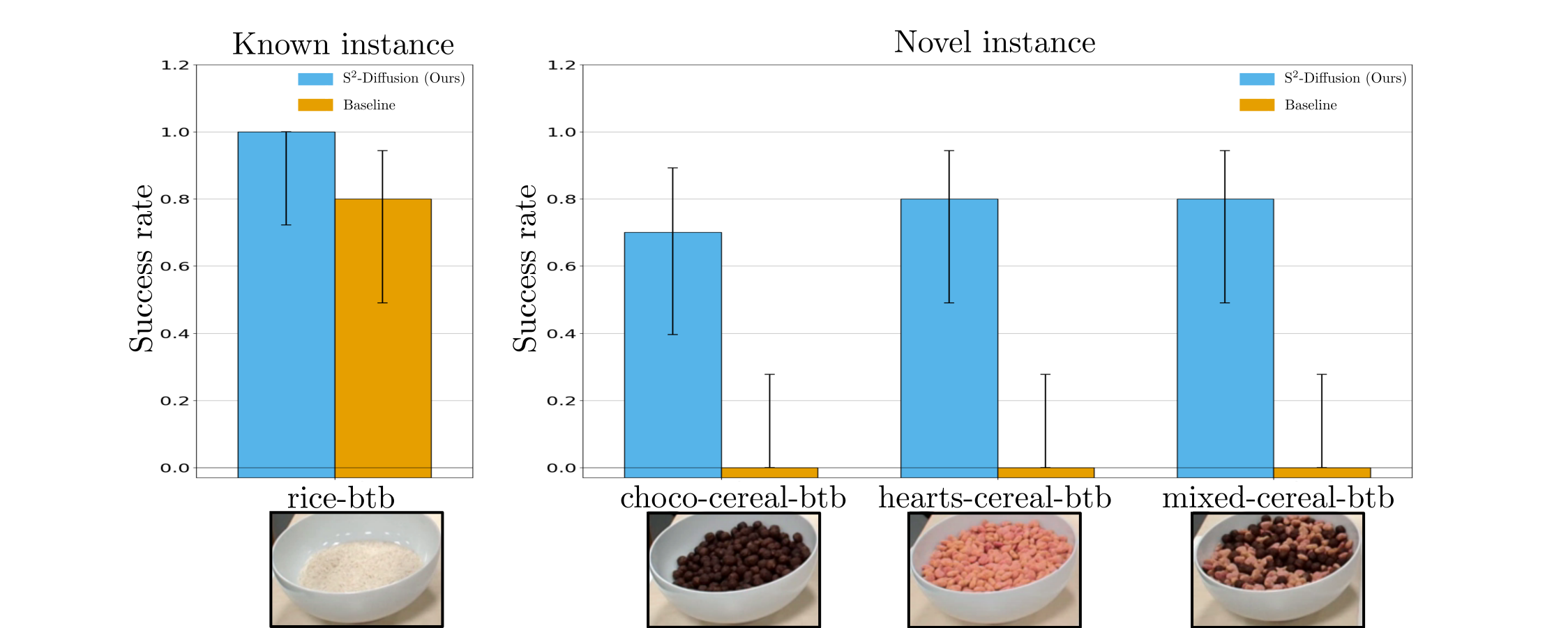}
        \label{fig:real scooping}
    }
    \vspace{-0.3cm}
    \caption{Comparison of our S$^2$-Diffusion and the baseline on two real-world environments: whiteboard wiping and bowl-to-bowl scooping. S$^2$-Diffusion and the baseline are trained on \textit{red-whiteboard-wiping} dataset and \textit{rice-bowl-to-bowl-scooping} dataset respectively, then evaluated on the known instances and transferred to unseen instances of the two tasks. Note that for \textit{choco-cereal-btb-scooping}, \textit{hearts-cereal-btb-scooping}, \textit{mixed-cereal-btb-scooping}, and \textit{green-whiteboard-wiping} the baseline diffusion policy shows $0\%$ success rate. }
    \label{fig: real-world results}
    \vspace{-\baselineskip}
\end{figure*}

\subsection{Real-World Experiments}

\noindent
\textbf{Experiment setup.}
We evaluate our S$^2$-Diffusion method on two real-world category tasks: \textit{whiteboard-wiping} and \textit{bowl-to-bowl scooping}. We collected $40$ and $60$ expert demonstrations for the red-whiteboard-wiping and rice-bowl-to-bowl scooping tasks instances respectively. The demonstrations were obtained by teleoperating a $7$-DOF Franka Panda manipulator using the Quest2ROS~\cite{welle2024quest2ros} Oculus app.
A single camera providing RGB observation was mounted on the end-effector as well as task-specific tools - such as a sponge for the whiteboard wiping and a spoon for the scooping task.
The language prompts for obtaining the semantic observations were \textit{"handwriting. sponge."} for the whiteboard wiping task and \textit{"rice. bowl."} for the bowl-to-bowl scooping task.


\noindent
\textbf{Baselines and Ablations.} 
We compare our S$^2$-Diffusion method with the visuomotor diffusion method from~\cite{chi2023diffusionpolicy} which uses RGB and proprioceptive information as a baseline, as well as perform a number of ablations on the scooping task - namely a version of our method that only has access to semantic observations (Semantic-Diffusion) and one that only observes the spatial observations (Spatial-Diffusion). Finally, we evaluate generalization to transparent objects.

\noindent
\textbf{Real-World Results.}
The real-world experimental results are reported in Fig.~\ref{fig: real-world results}.
We evaluate our method and the baseline first on the seen environment (red-whiteboard-wiping and rice-bowl-to-bowl-scooping) and evaluate the transferability on other instances of the same category - namely
black-whiteboard- and green-whiteboard-wiping and choco-cereal-, hearts-cereal-, mixed-cereal-bowl-to-bowl-scooping.

\noindent
\textbf{Whiteboard-wiping.} The task involves wiping scribbles of different colors (red, black, green) from a whiteboard using the end-effector-mounted wet sponge.
The results are shown in Fig.~\ref{fig:real wiping} including the $95\%$ Wilson confidence interval.
We define success as the robot completely removing the targeted handwriting using a sponge within $15$ seconds.
For the task instances that are covered by the training data (red-whiteboard-wiping), the baseline is able to succeed in all nine trials, leading to a $100\%$ success rate, however, once the policy is deployed on different instances of the task such as black-whiteboard-wiping the performance deteriorates to $4/9$ ($44\%$) and for the green-whiteboard-wiping task even to $0\%$. 
This clearly shows how even small changes in the RGB observations such as changing the color of the scribbles can lead to great deterioration and even complete failure of the skill on this instance - underscoring that the skill learning was indeed on the instance-level and \textbf{not} category-level. On the other hand, our S$^2$-Diffusion method has consistent performances across all tasks. These results highlight the ability of our S$^2$-Diffusion to learn category-level skills for handling novel tasks without requiring additional training or fine-tuning, a capability that is essential for real-world robotic applications. Note that the results of black-whiteboard-wiping are not statistically significant - we attribute this to the fact that the black observations are more similar to red than the green scribbles.

\noindent
\textbf{Bowl-to-bowl-scooping.}
The scooping task is to scoop granular materials from one bowl to another. The amount of material successfully scooped into the target bowl is measured in grams. Task success is defined as the policy's ability to scoop at least $3$ grams of material into the target bowl in under $30$ seconds.
In Fig.~\ref{fig:real scooping} the success rates are reported over $10$ trials. All real-world experimental videos are available on the project website.

\begin{figure*}[t!]
    \centering
    \vspace{-0.0mm}
    \includegraphics[width=1.0\linewidth]{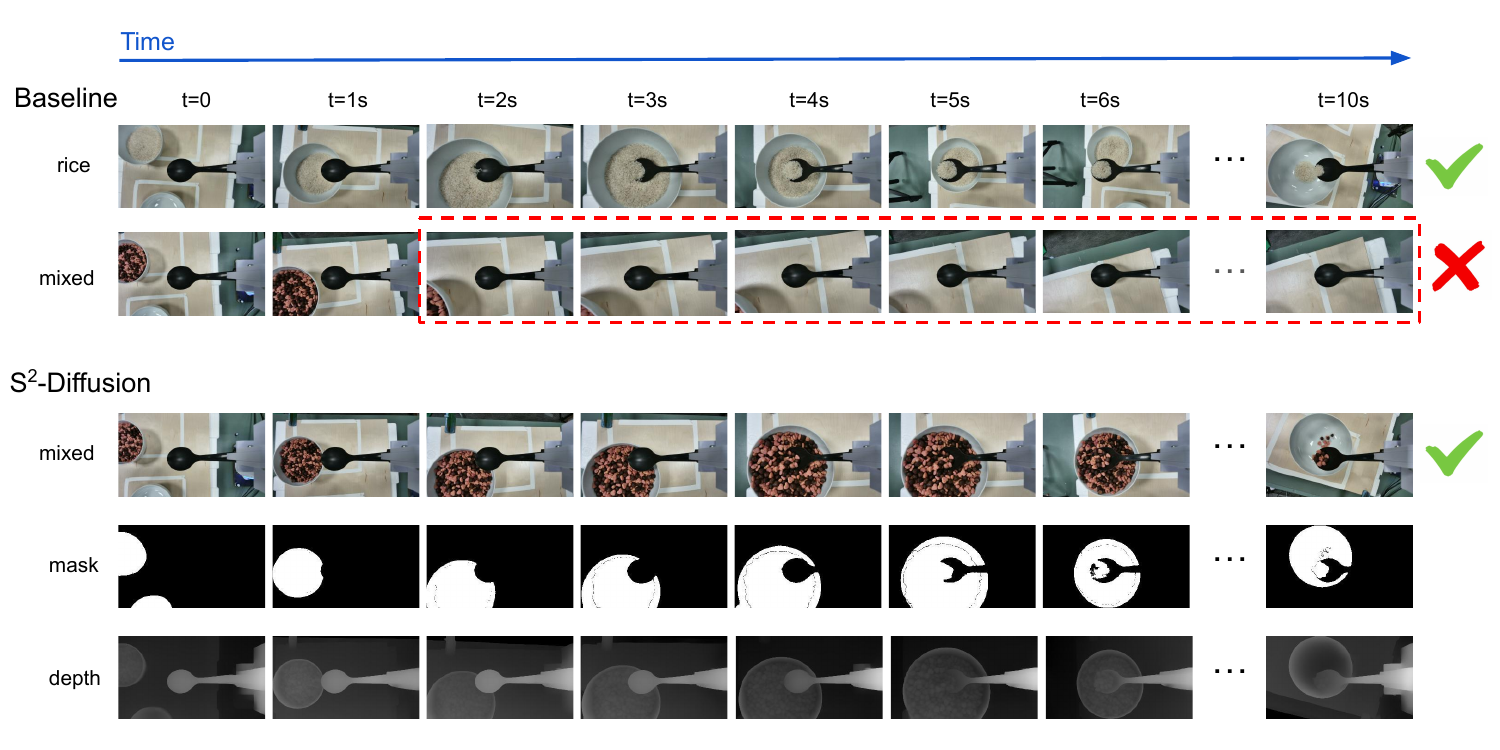}
    \vspace{-0.75cm}
      \caption{Frames of the baseline diffusion policy and our S$^2$-Diffusion method for real-world scooping tasks. We show the corresponding semantic mask and synthetic depth for each frame. The baseline can only succeed on the instance-level task it was trained on. Our method succeeds in learning a category-level skill trained on individual instances.
      }
      \vspace{-0.0mm}
      \vspace{-\baselineskip}
      \label{fig:real-s2}
\end{figure*}

For the in-distribution rice-bowl-to-bowl-scooping task, our S$^2$-Diffusion and baseline achieve success rates of $1.0$ and $0.8$, respectively. However, when transferring the policy to three unseen tasks—--choco-cereal-btb-scooping, hearts-cereal-btb-scooping, and mixed-cereal-btb-scooping—--the baseline diffusion policy fails entirely, with $0.0$ success rate, while our S$^2$-Diffusion policy maintains a high success rate of approximately $0.8$ by changing the semantic prompt to \textit{"cereal. bowl."} for all three instances. This demonstrates that our method effectively generalizes from individual instances to unseen other instances of the same category.
We can see frames of the baseline diffusion policy and our method in Fig.~\ref{fig:real-s2}, where the baseline can only succeed on the instance-level task it was trained on and fails to scoop out cereal as the visual observations are too different. Our method on the other hand extracts the semantic mask from the prompt as well as the spatial information via the synthetic depth observation. This leads to the successful execution of the scooping task and to the learning of a successful category-level skill trained on individual instances only using a single RGB observation (same as the baseline) as the original input.

\subsection{Ablation Evaluation}
\label{ablation evaluation}

\begin{figure}[t]
	\centering
	\includegraphics[width=\linewidth]{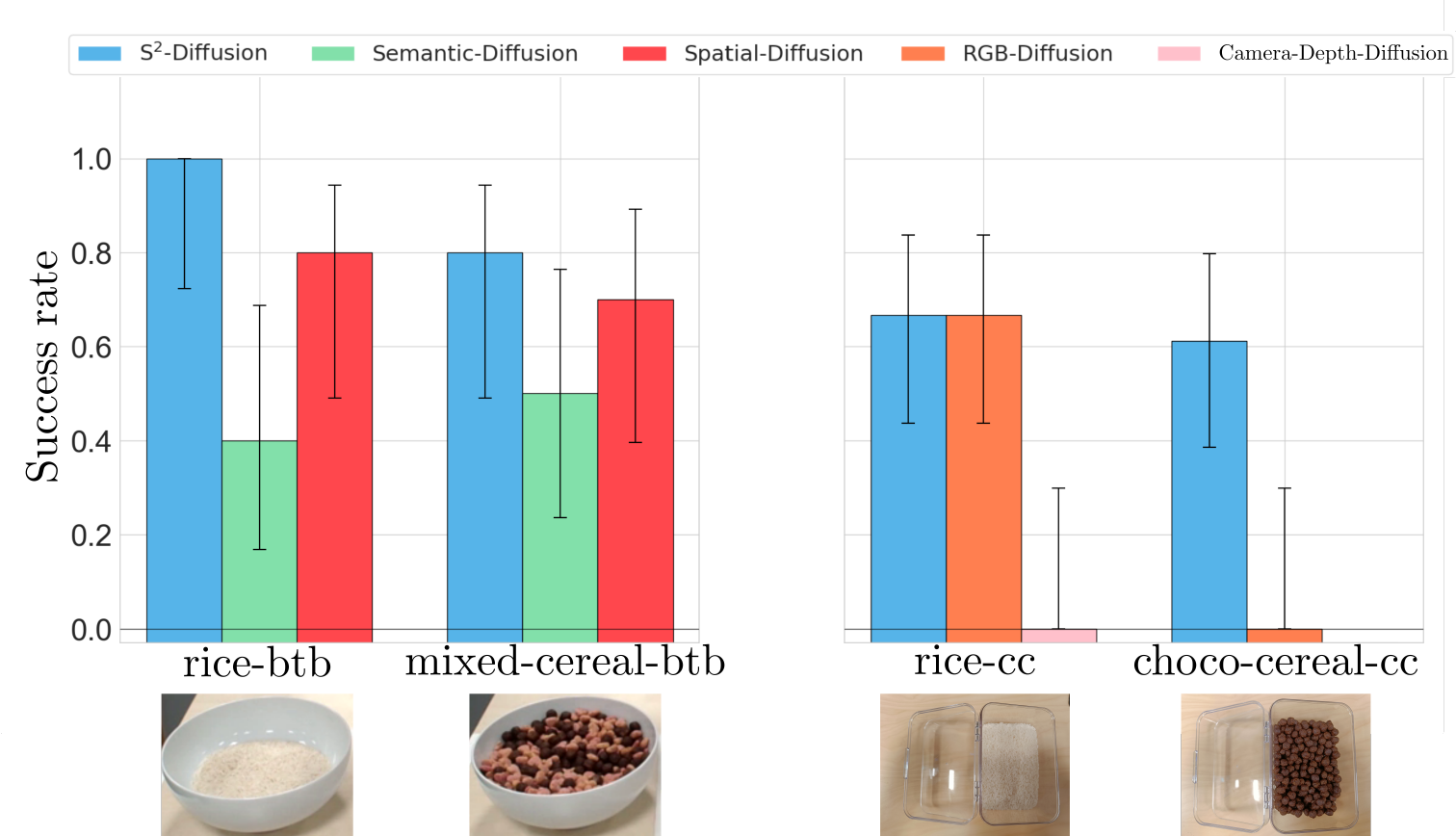}
	\vspace{-0.5cm}
	\caption{Ablation study of Semantic and Spatial-diffusion only on a seen task instance (\textbf{left}) and an unseen task instance (\textbf{right}). 
    S$^2$-Diffusion outperforms models trained with only semantic or spatial observations.
    }
	\label{fig:ablation only mask only depth} 
	\vspace{-0.3cm}
    \vspace{-\baselineskip}
\end{figure}

To investigate the significance of integrating spatial-semantic representation, we conduct additional experiments by training the policy using only the semantic representation (Semantic-Diffusion) and only the depth map estimation (Spatial-Diffusion) on the rice and mixed-cereal-bowl-to-bowl-scooping tasks. The results of these experiments are presented in Fig.~\ref{fig:ablation only mask only depth}. The full model S$^2$-Diffusion, which integrates both modalities, achieves the highest success rates, significantly outperforming the Semantic-Diffusion and also etching out the Spatial-Diffusion versions both on the seen task and also the novel task.
When the depth information is excluded, the performance of the policy drops drastically, with success rates falling to $0.4$ and $0.5$ for the seen and novel tasks, respectively. This significant decline underscores the critical role that spatial information plays in successfully executing 3D tasks. 
To evaluate generalization to transparent objects, we add a task where it has to close a transparent container. For reference, we also compare against a DP with depth camera input, trained directly on the transparent container. We find that S$^2$ Diffusion only has a small reduction in success rate while the depth camera policy was not able to complete the task - illustrating the robustness of the S$^2$ approach in common kitchen tasks involving transparent or reflective surfaces - where depth sensors struggle.
In contrast, when the mask (semantic representation) is removed, the performance also declines, but to a lesser extent compared to the removal of depth information. While the policy's ability to generalize across tasks is still reasonable in this case, it benefits from the combined use of both modalities.

\section{Limitations}

As Fig.~\ref{fig:skill_h} illustrates, functional goals like \textit{flipping} or \textit{scooping} can be discretized into different category-level skills such as \textit{bowl-to-bowl} and \textit{pile-to-container} scooping. While our method generalizes from instance-level to category-level tasks, it does not generalize well across different categories; for example, a policy trained on \textit{rice-bowl-to-bowl} scooping struggles with a \textit{sand-pile-to-container} task.
Although the abstraction hierarchy in Fig.~\ref{fig:skill_h} appears straightforward, real-world functional goals can depend on more complex context 
and categories like \textit{bowl-to-bowl scooping} may require additional qualifiers. For example, transferring frozen ice cream differs significantly from transferring cereals due to the forces required, although generalization may improve as the ice cream melts. Addressing such nuanced skill ontologies is an avenue for future work.
Another limitation is the dependency on the performance of pretrained VLMs as well as proper text prompts for accurate semantic segmentation. Poor semantic or spatial estimates can degrade category-level skill generalization, particularly for greatly different object instances. Fine-tuning VLMs jointly with the policy network is a promising direction for mitigating this dependency.

\section{\uppercase{Conclusion}}
\label{sec:conclusion}

We introduced S$^2$-Diffusion, an open-vocabulary spatial-semantic diffusion policy that enables generalization from instance-level training data to category-level skills in robotic manipulation. By integrating semantic understanding and spatial representations via vision foundation models, our method learns policies invariant to task-irrelevant visual changes, allowing skill transfer without additional fine-tuning.
Through extensive simulations and real-world evaluations, we demonstrated that S$^2$-Diffusion outperforms baselines. In particular:
\begin{itemize}
    \item \textbf{Spatial-Semantic representations enhance generalization:} By combining spatial and semantic information, our method focuses on task-relevant features, enabling instance-to-category transfer.
    \item \textbf{Efficient real-time execution:} S$^2$-Diffusion requires only a single RGB camera, avoiding multi-view setups or depth sensors, making it practical for real-world deployment and attractive for mobile settings.
    \item \textbf{Category-level generalization:} Our evaluations show strong generalization across unseen instances, achieving high performance where baseline policies fail.
\end{itemize}
Overall, S$^2$-Diffusion represents a step toward enabling robots to generalize skills across variations in objects, materials, and environments, similar to human capabilities.

\bibliographystyle{IEEEtran}   
\bibliography{references}




\end{document}